\documentclass[11pt]{article}

\usepackage[preprint]{acl}

\usepackage{times}
\usepackage{latexsym}

\usepackage[T1]{fontenc}

\usepackage[utf8]{inputenc}

\usepackage{microtype}

\usepackage{inconsolata}

\usepackage[dvipsnames]{xcolor}
\usepackage{graphicx}
\usepackage{enumitem}
\usepackage{xurl}

\usepackage{booktabs}
\usepackage{tabularx}
\usepackage{amssymb}
\usepackage{siunitx}
\usepackage{comment}
\usepackage{makecell}
\usepackage[table]{xcolor}

\usepackage[most]{tcolorbox}
\tcbset{
  promptbox/.style={
    colback=blue!5,
    colframe=black!60,
    colbacktitle=gray!20,
    coltitle=black,
    fonttitle=\bfseries,
    title={#1},
    fontupper=\ttfamily\small,
    boxrule=0.8pt,
    arc=2mm,
    boxsep=2pt,
    left=6pt,
    right=6pt,
    top=4pt,
    bottom=4pt,
    enhanced
  }
}

\definecolor{headerbg}{RGB}{45, 62, 80}
\definecolor{headerfg}{RGB}{255, 255, 255}
\definecolor{breakline}{RGB}{200, 200, 200}

\newtcolorbox{ragcomparison}[1][]{
    enhanced,
    colback=white,
    colframe=headerbg,
    fonttitle=\bfseries\large,
    coltitle=headerfg,
    colbacktitle=headerbg,
    attach boxed title to top center={yshift=-2mm},
    boxed title style={
        colback=headerbg,
        sharp corners,
        boxrule=0pt,
        left=10pt,
        right=10pt,
        top=5pt,
        bottom=5pt
    },
    sharp corners,
    boxrule=1pt,
    top=8mm,
    bottom=5mm,
    left=5mm,
    right=5mm,
    #1
}

\newcommand{\methodsection}[2]{%
    {\centering\colorbox{headerbg}{\textcolor{headerfg}{\bfseries\strut\ #1\ }}\par}
    \vspace{2mm}
    #2
}

\newcommand{\brokenline}{%
    \vspace{3mm}
    \noindent\makebox[\linewidth]{\textcolor{breakline}{%
        \hspace{-5mm}\rule{0.45\linewidth}{0.5pt}%
        \hspace{2mm}\raisebox{-1pt}{$\cdots$}\hspace{2mm}%
        \rule{0.45\linewidth}{0.5pt}\hspace{-5mm}%
    }}
    \vspace{3mm}
}

\definecolor{reasoningbg}{HTML}{F8F9FA}     
\definecolor{reasoningline}{HTML}{6C757D}   
\definecolor{reasoningtext}{HTML}{495057}   
\definecolor{reasoningtitle}{HTML}{343A40}  

\newtcolorbox{llmreasoning}[1][]{
    enhanced,
    boxrule=0pt,
    frame hidden,
    sharp corners,
    colback=reasoningbg,
    coltext=reasoningtext,
    borderline west={3pt}{0pt}{reasoningline}, 
    fonttitle=\bfseries\color{reasoningtitle},
    fontupper=\small\itshape, 
    attach title to upper,
    after title={\par\medskip},
    left=12pt, 
    right=12pt,
    top=8pt,
    bottom=8pt,
    before skip=12pt,
    after skip=12pt,
    #1 
}

\newcommand{\decr}[1]{\rlap{\textbf{\textsuperscript{\textcolor{red}{\fontsize{5.5}{0}\selectfont \,#1}}}}}

\newcommand{\suhang}[1]{\textcolor{blue}{[SW: #1]}}

\title{\textit{Can Hallucinations Be Useful?} Solving Multi-Hop Questions With SLMs By Chaining System-I/II Reasoning}

\author{Saptarshi Sengupta \\
    The Pennsylvania State University \\
    \texttt{sks6765@psu.edu} \\
  \\\And
  Suhang Wang \\
  The Pennsylvania State University \\
  \texttt{szw494@psu.edu} \\
}

\begin{document}
\maketitle
\begin{abstract}
Recently, there has been increased interest in Small Language Models (SLMs), which are fast, show good performance, and have lower hardware demands than large language models (LLMs). However, SLMs hallucinate more frequently than LLMs, impacting their ability to solve complex multi-step reasoning problems as early mistakes cascade to the final response. To address this, existing works \textit{think-first} followed by iterative retrieval to reduce hallucination. We argue that the \textit{think-first} strategy is not always necessary as we find that: (i) SLMs are often accurately confident in their initial answer and, (ii) \textit{hallucinations can actually be beneficial} for honing in on the true answer. As such, we position our work as an \textit{inversion} of this strategy, i.e., \textit{answer first-reason later}. We propose a \textit{cognitively-inspired} framework where the model is first allowed to quickly answer the question (System-I (zero-shot)) and then resorts to deeper thinking (System-II) based on evidence retrieved from a knowledge source using the initial hypothesis. By combining System-I and System-II style thinking, we show that our method can outperform prior work that takes the traditional think-first route on various multi-step question-answering benchmarks.
\end{abstract}

\section{Introduction}

Multi-hop (or step) question answering (MHQA) requires a model to chain multiple pieces (hops) of information to answer a question \cite{9960856}. For example, the question ``\textit{Who was the top goal scorer in the first season of the team that won back-to-back Champions League titles in the last 10 years?}'' needs a model to (i) find out which teams won in the last 10 years, (ii) determine which team won back-to-back, and (iii) find the first season's top scorer. Large Language Models (LLMs) \cite{10.1145/3744746} have demonstrated substantial potential in solving such questions owing to their strong \textit{reasoning} abilities \cite{XU2025101370}. However, deep reasoning is generally observed in truly large models such as GPT-5 \cite{GPT-5} and K2 \cite{team2025kimi}. Developing such models is non-trivial, making their wider adoption difficult.



Recently, there has been an increasing interest in \textit{small} language models (SLMs) \cite{10.1145/3711896.3736563}, which offer attractive properties, e.g., lower hardware demands and reasonably good performance. However, SLMs can hallucinate (plausible but incorrect response) \cite{10.1145/3703155} more frequently than LLMs \cite{li-etal-2024-dawn}. This hurts performance as early mistakes might propagate through reasoning, leading to incorrect answers.

An obvious solution to mitigate hallucinations is to train SLMs with specialized data \cite{li2024small, tian2024finetuning}. However, as acquiring high-quality training data is non-trivial, research \cite{srivastava-etal-2025-thinkslm} has shifted to more efficient reasoning techniques. Test-Time Scaling (TTS) \cite{zhang2025survey} is one such technique that has gained recent popularity. It involves allocating more compute during inference by promoting models to generate long chains-of-thought (CoT) \cite{wei2022chain} before answering. TTS is generally applied to \textit{logic} problems, such as math or coding \cite{nvidiaEasyIntroduction}, which \textit{do not require external information} and can be solved by a model using only its parametric knowledge.

Although TTS has shown promise for logic problems, recent work \cite{DBLP:journals/corr/abs-2509-06861} suggests that it is infeasible for knowledge-heavy tasks and can even exacerbate hallucinations. This is because a model cannot reason its way into an answer if it fundamentally lacks background information on a topic. To alleviate this, they are coupled with external knowledge via retrieval-augmented generation (RAG) \cite{DBLP:journals/corr/abs-2312-10997}. In this work, we view advanced RAG, i.e., beyond one-shot retrieve-then-generate, as a form of \texttt{knowledge-based TTS} (KB-TTS) where scaling happens across reasoning tokens (depth) and a knowledge source (breadth). 

\begin{figure}
    \centering
    \includegraphics[width=\columnwidth]{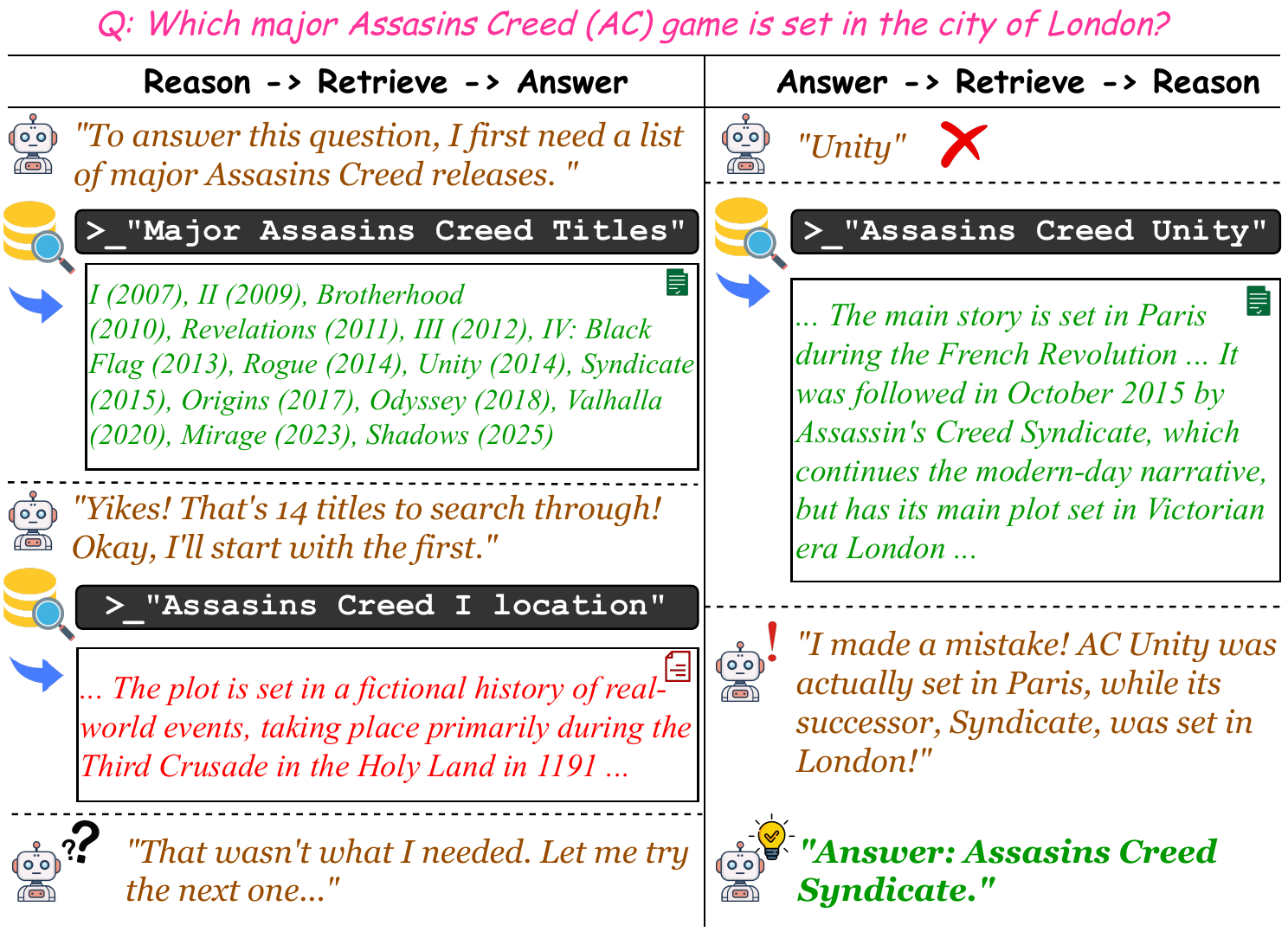}
    \vskip -0.5em
    \caption{Current methods alternate between reasoning and retrieval, leading to the model getting flooded with unnecessary information, which ultimately misses the answer. Our method asks a model to give a fast, but might be wrong answer, which serves as an anchor to locate evidence for/against it, leading to the true answer.}
    \label{fig:motivation}
    \vspace{-1em}
\end{figure}

Most KB-TTS approaches \cite{li-etal-2025-survey} follow a \textit{plan-retrieve-reason-answer} cycle for MHQA, i.e., iteratively decomposing a question, retrieving background information, verifying the correctness of each step, and finally generating the answer. This is done to prevent a model from hallucinating as much as possible by breaking down a question into simpler and more manageable sub-questions. While this works, we argue that it is not always necessary for three reasons: (i) In cases where the model is accurately confident in itself. For example, \textit{without thinking}, an SLM can accurately answer questions like ``\textit{Which continent is the Eiffel Tower located in?}''. Iteratively solving questions like this is costly whereas an early stop would be favorable; (ii) Due to their size, SLMs struggle to reason at the level of LLMs as their breakdown of queries may be imperfect \cite{kim2025guiding, li-etal-2025-small-models}; (iii) When the model cannot find the needed information, it spirals into a chain of retrievals that continue to grow its context window (c.f. Table \ref{tab:token_comparison}), finally resulting in noisy contexts and wrong answer. For example, in Fig. \ref{fig:motivation} (left), the SLM correctly identifies the search target (c.f. App. \ref{sec:app_slm_reasoning} for empirical validation). However, if the queries do not return anything useful, the model will keep exploring till either (i) it exhausts its context window or (ii) it exceeds the maximum requests allowed by the framework.

To address this, we suggest an \textit{inversion} of the paradigm, i.e., instead of think-first, we propose \textit{answer-first-reason-later}, framing our approach as cognitively inspired System-I (zero-shot: fast/instinctive) and System-II (post evidence gathering: slow/deliberate) type thinking~\cite{kahneman2011thinking}. We observe this with many human experiences. For example, when a physician examines a complicated case, they may first offer a \textit{gut-feel} hypothesis (System-I) based on the immediate information. However, to confirm it (System-II), they perform tests, study patient history and consider plausible alternatives \cite{rotgans2015time}. This dual-thinking model informs the design of our approach.

To this end, given a question, we first ask an SLM to answer quickly, without thinking. We notice that even if the answer is wrong, there exists some element of truth to its rationale. For instance, in Fig. \ref{fig:motivation} (right), with System-I thinking, an SLM might answer \textit{Unity} (App. \ref{sec:app_slm_reasoning}). However, \textit{Unity} is set in \textit{Paris} \cite{enwiki:1354537151}, while its successor, \textit{Syndicate}, is set in \textit{London}. The model learns this by studying the initial answer's background. If we imagine the context as a graph, this mistake is plausible as the initial answer is a neighbor of the true answer. The observation here is that while the initial answer is incorrect, it is in the vicinity of the truth, leading us to argue that \textit{hallucinations might not always be detrimental} and, on the contrary, can provide an initial clue leading to the correct answer. We use this insight for System-II, where we treat the initial answer as a \textit{hypothesis} to initialize reasoning. Here, the SLM is first asked to generate knowledge-graph style triples (subject-predicate-object) from the initial answer and question (Fig. \ref{fig:framework}). These triples, containing entities and relationships relevant to the question (\S \ref{sec:triple_gen}), are considered as a condensed summary of the SLM's reasoning and would be useful for retrieval. We use the triples as anchors to search a knowledge base for supporting or counterfactual contexts, if the initial answer is incorrect. The SLM studies all of the information, identifies its mistakes, and finally provides a refined answer.

Overall, our \textbf{contributions} are: (i) We identify the utility of hallucinations as an effective starting point in searching for information related to a query; (ii) We challenge the existing paradigm of \textit{reason-first} KB-TTS with a simple and unique \textit{answer-first-reason-later} strategy; (iii) Results indicate the effectiveness of our approach against more sophisticated approaches that generate detailed reasoning traces and prune invalid trajectories.

\section{Related Work}
\label{sec:related_work}

To ensure comprehensiveness, we discuss training-free and training-based approaches to MHQA.

\textbf{SFT on Distilled LLM CoT}: Generally, LLMs craft superior reasoning traces (CoT) than SLMs. Observing this, prior work \cite{kang2025distilling, bi-etal-2025-enhancing, lee-etal-2025-steper, piao-park-2025-tinythinker} distills synthetic reasoning data from LLMs to train SLMs to mimic them through supervised fine-tuning (SFT). However, as pointed out by \citet{gudibande2024the}, synthetic CoT makes SLMs mimic an LLM's \textit{style}, but not their \textit{factuality}. This means that a trained SLM CoT follows similar steps as an LLM to break down a problem, but its core content is factually incorrect. 

\textbf{Reinforcement Learning (RL)}: As shown by \cite{yang2025qwen3, guo2025deepseek}, SFT followed by RL builds good reasoning as it encourages creative exploration of a solution space. Indeed, this has shown promise in various MHQA tasks \cite{yu2025graphrag, luo2025graph, R1-searcher}. A critical RL requirement is the design of proper reward functions to guide the learned policy (trained SLM). Unfortunately, none of the mentioned works employ rewards to check \textit{what} is being retrieved and only rely on surface-level response format checks. Even when they do evaluate retrieval quality \cite{luo2026d}, the checks are imperfect, leading to degraded system performance. Considering the importance of retrieved content in RAG \cite{wu-etal-2025-pandoras}, this class of methods appears sub-optimal for MHQA.

\textbf{Test-Time Scaling}: As an alternative to training, TTS explores techniques to improve model performance at inference time by having it \textit{think} harder (generate reasoning tokens) before answering. Early attempts at TTS include Tree  \cite{NEURIPS2023_271db992}, Graph \cite{graphofthoughts}, and Forest \cite{bi2025forestofthought} - of Thoughts, all building on the idea of exploring multiple reasoning trajectories before converging on the most promising one. These methods were developed for math or logical reasoning tasks, i.e., not requiring external information. For non-math/logic tasks, exhaustive reasoning cannot bridge the gap between knowing and not-knowing information on a given entity \cite{DBLP:journals/corr/abs-2509-06861}.

KB-TTS methods \cite{pan-etal-2025-coat, liu2026graph, luo2025hypergraphrag, jiao2026prunerag, shi2026reasoning, li2026structured} address the above limitation. Here, a retriever provides the necessary context for a query in line with the model's reasoning. These methods are typically \textit{reason and retrieve}, i.e., iteratively break down a question into sub-questions and retrieve context to answer them. However, progressive question solving relies on strong decomposition abilities, which are not always reliable for SLMs \cite{kim2025guiding, li-etal-2025-small-models}. 

We propose an inversion of this strategy, i.e., instead of reasoning first, we use an SLM's initial guess to guide evidence acquisition used to inform its final response. Our approach hinges on the observation that an SLM's initial answer, although potentially hallucinated, is often in the vicinity of the ground truth. We leverage this to zero in on the true answer, instead of exploring dead-end reasoning paths. To the best of our knowledge, we are the first to make use of hallucinations in the proposed manner. The only other related work we find is \citet{hosseini2024vstar}, who \textbf{trains} a reward model based on an LLM's correct/incorrect trajectories to better inform its decision making. While our methods share the same spirit, they are ultimately orthogonal as we approach MHQA through a \textbf{training-free} framework.

\section{Methodology}

\begin{figure*}
    \centering
    \includegraphics[width=0.8\linewidth]{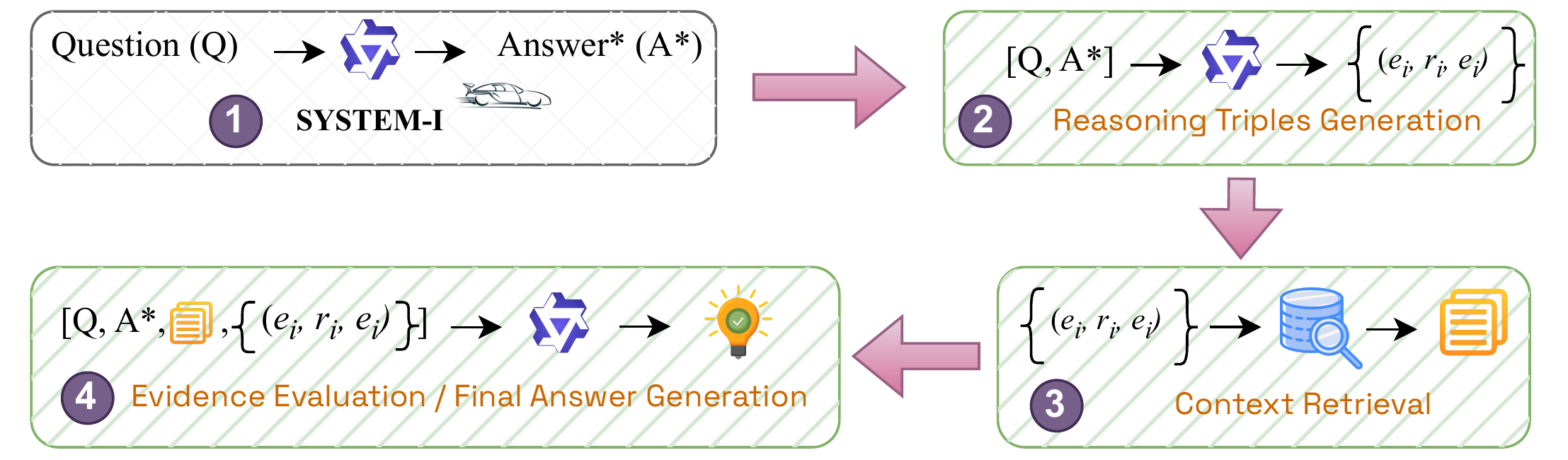}
    \vskip -0.5em
    \caption{Method Overview. Step 1 (\textcolor{gray}{gray} provides System-I answer). Steps 2-4 (\textcolor{ForestGreen}{green}) involve System-II operations.}
    \label{fig:framework}
    \vspace{-5pt}
\end{figure*}


An overview of our method is shown in Figure \ref{fig:framework}. At a high-level: given a question, the SLM first generates a quick, initial answer, serving as its hypothesis (System-I). Using this, it produces a condensed reasoning trace in the form of knowledge-graph style triples, which are used as search queries to retrieve context that may support or refute the initial claim. Finally, the SLM deliberates over the initial answer, triples, and retrieved context to identify its mistakes and come up with the final correct answer (System-II).

\subsection{System-I: Direct Answer}

Our framework begins by simply having the SLM try its best to provide an answer to the question by leveraging its parameterized knowledge. Here, the model is deliberately instructed not to reason (\textbf{no CoT}) before answering. Formally, given a question $Q$, the SLM generates an initial answer $A^*$ following prompt $P_{Sys-1}$ ( in Fig. \ref{fig:sys1_prompt}) as
\begin{equation}
   \label{eq:s1}
   A^* = SLM(Q,P_{Sys-I})
\end{equation}

The goal in this step is to harness the SLMs \textit{parametric knowledge} to guide further evidence acquisition. As noted previously, an SLM is often capable of either producing a correct or near-correct answer to a question. We use this insight here to narrow the search space of a question instead of having it reason at the onset, which has a potentially higher degree of being error-prone due to their weaker question-decomposition abilities (\S \ref{sec:related_work}).

\subsection{System-II: Reasoning With Evidence}

System-II requires a more deliberate and well-rationed thought process to answer a question. As such, the \textbf{bulk of the reasoning budget is spent here}. This phase consists of three steps: (i) \textit{Reasoning Triple Generation}: Given the initial answer, the SLM explains, using knowledge-graph style triples how it arrived at the answer; (ii) \textit{Evidence Collection}: The reasoning triples are used as anchors to locate context relevant to the question from a knowledge base; (iii) \textit{Final Answer}: The SLM reviews everything, investigates if/where there are flaws are and forms its final judgment. 


\subsubsection{Reasoning Triple Generation}
\label{sec:triple_gen}

System-II starts with the SLM generating a set of \textit{reasoning triples}. Given the question ($Q$) and its System-I answer ($A^*$), the model is asked to explain how it arrived at the answer in the form of (subject-predicate-object) triplets using the prompt $P_{triple\_gen}$ (in Fig. \ref{fig:triple_gen_prompt}), which can be written as
\begin{equation}
    \label{eq:triple_gen}
    \{(e_i, r_i, e_i)\} = SLM(Q, A^*, P_{triple\_gen})
\end{equation}
where $e_i$ denotes entity and, $r_i$ the relationship between the two entities. This is done for two reasons: (i) when explaining multi-hop logic, we view triples as a form of condensed reasoning that is easier for the SLM to handle than a verbose explanation; and (ii) each triple acts as a search key for locating a particular slice of information relevant to the question (\S \ref{sec:context_retrieval}). 

We realize that the triples might contain hallucinated entities, as the model is expected to form connections using only its parametric knowledge. For example, given the question \textit{What league does Luka Modric play in?}, an SLM might answer \textcolor{red}{\textit{La Liga}} by yielding triples as, \textit{Modric | plays\_for | AC Milan} and \textit{AC Milan | competes\_in | \textcolor{red}{La Liga}} (correct - \textcolor{ForestGreen}{\textit{Serie A}}). Such mistakes are plausible due to a model's knowledge cutoff \cite{gao-etal-2025-prompts} or popularity biases \cite{10.1145/3774904.3792168}. Even though the second triple is incorrect, it contains relevant information, i.e., the concerned team, that can help a retriever perform \textit{approximate neighbor matching} to locate related context. We reason that, as long as the key entities in the triples are related to the question, there is a high chance of acquiring the correct or, at least, connected information. 

\subsubsection{Context Retrieval}
\label{sec:context_retrieval}

Each generated triple describes a particular relationship associated with a question. They act as search keys to locate context relevant to their highlighted relationship. For example, the triple \textit{Modric | plays\_for | AC Milan} can be used to retrieve chunks related to the player and his team. The retrieved context then serves as \textbf{grounding evidence} which may support or refute the System-I answer, depending on whether it was correct. This context is necessary for the SLM in the last step (\S \ref{sec:final_gen}) where it must consider all of the provided information to reach its final decision.
Specifically, given a reasoning triple, generated in \S \ref{sec:triple_gen}, a $Retriever$ fetches the top-k relevant document chunks from the provided knowledge base as
\begin{equation}
    \label{eq:retriever}
    D = Retriever(\{(e_i, r_i, e_i)\})
\end{equation}
where $D = \{d_i\}_{i=1}^{k}$ is the set of retrieved documents. Implementation-wise, the retriever performs simple semantic similarity (via approximate neighbor matching) between the search queries (triples) and the knowledge index. Although a triple is expressed in a pseudo-language (e.g., \textit{plays\textunderscore in}), a relatively strong retriever is capable of distilling the underlying meaning of the phrase. For each triple, we retrieve their top-5 document chunks and unify the results by removing repeated entries.

\subsubsection{Final Generation}
\label{sec:final_gen}


After all of the evidence has been gathered, the SLM is in a position to make an informed judgment about the question. Formally, the SLM utilizes the question ($Q$), System-I answer ($A^*$), generated triples ($\{(e_i, r_i, e_i)^R\}$) to support that answer (\S \ref{sec:triple_gen}) and the context retrieved ($D$) using the triples (\S \ref{sec:context_retrieval}) to form its final answer ($\mathcal{A}$) as
\begin{equation}
    \label{eq:final_gen}
    \mathcal{A} = SLM(Q,\! A^*,\! D,\! \{(e_i,\! r_i,\! e_i)\},\! P_{Sys-II})
\end{equation}
where $P_{Sys-II}$ is the prompt used (see Fig.\ref{fig:sys2_prompt}). We provide the initial answer and reasoning trace alongside the new evidence to the model to give it a bird's-eye view of all the data it needs to provide the final answer. With these, the SLM can (i) contrast retrieved context with its initial line of reasoning (triples) that led it to the System-I answer; (ii) identify potential flaws in its initial reasoning; (iii) update its rationale in light of the retrieved evidence, and (iv) provide a final System-II answer to the question. As reasoning is naturally critical to this step, we use a higher generation limit to promote the model to think more before answering. 

\section{Experiments}

Through our experiments, we aim to answer three research questions (RQ) - (i) How well does our method perform against prior work? (ii) What is the overall impact of System-I v/s II answering? and (iii) What does a qualitative analysis of the framework reveal about the success and failure modes?

\subsection{Experiment Setup}
\label{sec:expertiment_setup}

We first describe the datasets, models and baselines used in our work. 


\textbf{Datasets:} We evaluate our framework on three widely used MHQA datasets, i.e., 2WikiMultiHopQA \cite{ho-etal-2020-constructing}, HotpotQA \cite{yang-etal-2018-hotpotqa}, and MuSiQue \cite{trivedi-etal-2022-musique}. The questions in these datasets require a model to assemble information from connected contexts to arrive at a solution. Following \citet{jiao2026prunerag, liu2026graph}, we randomly sample 500 examples from each dataset's validation split for evaluation.

\textbf{Index:} To provide background for the questions, each dataset contains Wikipedia-sourced, abstract-style contexts, relevant or irrelevant to the question. Following prior work \cite{Wang_Deng_Guan_Lu_Jiang_2026, pan-etal-2025-coat}, we construct our knowledge retrieval index (\S \ref{sec:context_retrieval}) by combining all of these contexts, for each dataset. As the maximum context length ranges between \textasciitilde900 and 2K tokens, we create standard \cite{firecrawlBestChunking} 400 token chunks with a stride (context overlap) of 50 tokens for continuity.

\textbf{Models:} Following (most) prior work, we use \texttt{Qwen2.5-7B-Instruct} \cite{qwen2024qwen2} as the SLM for all our experiments. Qwen2.5 is not a native reasoning model, i.e., does not default to long-form thinking using \texttt{<think>} tags. Thus, it is ideal for isolating the strength of reasoning frameworks without the influence of pre-trained biases. For our retriever, we use a lightweight \texttt{Qwen3-0.6B} \cite{zhang2025qwen3} embedding model, which offers high retrieval accuracy and low-compute requirements.

\textbf{Baselines:} As our framework does not involve post-training of the SLM, we compare it against similar methods. For competitiveness, we select the latest methods published near the time of writing this paper. Unfortunately, as many papers, including \textit{several} prior published works, have not released code, we could not test them. That said, we tried our best to be fair by selecting the following categories of methods: (i) \textbf{One-shot retrieval} that performs context retrieval once based on the search query. This includes \textbf{Standard RAG} \cite{NEURIPS2020_6b493230}, i.e., retrieving top-5 document chunks for a question, and \textbf{HyperGraphRAG} \cite{luo2025hypergraphrag} that uses \textit{knowledge-hypergraphs}, i.e., graph structures where edges connect multiple vertices, as their knowledge index; (ii) \textbf{Iterative retrieval} that performs multiple rounds of retrieval based on the decomposed query. This includes \textbf{GraphAnchor} \cite{liu2026graph} which uses an \textit{evolving} knowledge-graph; \textbf{PruneRAG} \cite{jiao2026prunerag} and \textbf{RT-RAG} (Reasoning Tree Guided RAG) \cite{shi2026reasoning}, both of which explore a solution space by rejecting search-paths with low-quality answers; \textbf{PAGER} \cite{li2026structured} which uses \textit{placeholder clues} generated by decomposing the question to inform the retrieval context; and \textbf{Search-o1} \cite{li-etal-2025-search}, which weaves search and an evolving knowledge-store to guide an agent. Further details are provided in App. \ref{sec:app_baselines}.


\textbf{Metrics:} We report standard MHQA metrics (App. \ref{sec:app_eval_dets}): (i) Exact Match (EM), 0/1 measure of exact string match; (ii) F1, harmonic mean of token overlap between the predicted and gold answer. 

\subsection{RQ1: Performance Against Prior Work}

\begin{table*}[ht]
\centering
\small
\begin{tabular}{lcccccccc}
\toprule
 & \multicolumn{6}{c}{Dataset} \\
 \cmidrule(r){2-7}
Method & \multicolumn{2}{c}{2WikiMultiHopQA} & \multicolumn{2}{c}{HotPotQA} & \multicolumn{2}{c}{Musique} & \multicolumn{2}{c}{Average} \\
\cmidrule(lr){2-3} \cmidrule(lr){4-5} \cmidrule(lr){6-7} \cmidrule(lr){8-9}
 & EM & F1 & EM & F1 & EM & F1 & EM & F1 \\
\midrule
Standard RAG & 26.8	& 33.16 & 16.4 & 25.08 & 8.6 & 15.9 & 17.27 & 24.71 \\
HyperGraphRAG$\ast$ & 0.40 & 12.82 & 0.40 & 8.93 & 0.60 & 10.03 & 0.47 & 10.59 \\
GraphAnchor$\dagger$ & 22.4 & 28.23 & 22.2 & 29.44 & 9 & 14.62 & 17.87 & 24.1 \\
PruneRAG & 19.68 & 23.35 & 15.12 & 21.56 & 2.60 & 7.21 & 12.47 & 17.37 \\
RT-RAG & 15.40 & 22.05 & 9.40 & 14.89 & 6.20 & 12.33 & 10.33 & 16.42 \\
PAGER & 11.80 & 24.06 & 8.80 & 20.31 & 2.80 & 9.42 & 7.80 & 17.93 \\
Search-o1 & 19.4 & 26.44 & 12.8 & 22.3 & 4.8 & 12.22 & 12.33 & 20.32 \\
\midrule
\textbf{Ours (Proposed)} & \textbf{29.60} & \textbf{34.71} & \textbf{20.80} & \textbf{28.75} & \textbf{10.20} & \textbf{19.42} & \textbf{20.20} & \textbf{27.63} \\
\bottomrule
\end{tabular}

\vspace{-1.5mm}
\caption{Comparing frameworks. $\ast$Uses Gold Context only, $\dagger$Unable to complete \textasciitilde43\% questions (App. \ref{sec:app_baselines}).}
\label{tab:main_results}
\vspace{-10pt}
\end{table*}


We present our main results in Table \ref{tab:main_results}. As we can see, our proposed approach dominates existing \textit{reason-then-retrieve} strategies. This reinforces our hypothesis that smaller models struggle to generate cogent plans (\S \ref{sec:related_work}) and thus suffer in performance when subjected to such frameworks.


HyperGraphRAG and GraphAnchor are variations of GraphRAG \cite{10.1145/3777378}, i.e., use knowledge graphs instead of documents as their index. The former's poor performance highlights major limitations with GraphRAG approaches \cite{10.1145/3777378}, including the complexity of developing knowledge-graphs (App. \ref{sec:app_baselines}). Although they perform a one-time retrieval, they are ultimately let down by the complex nature of their retrieval structure (Table \ref{tab:token_comparison}). GraphAnchor augments our text index with an evolving graph index, used for relational information. However, the complexity of parallelly decomposing questions, building graphs, and reasoning about contexts, overloads an SLM, preventing it from even being able to process all questions (App. \ref{sec:app_baselines}). In comparison, we use simple text chunks, similar to traditional RAG \cite{DBLP:journals/corr/abs-2312-10997} and avoid complex operations such as weaving graph construction with question reasoning.

The remaining four methods utilize a text index and perform multiple rounds of reasoning and retrieval. Empirically, these approaches work better than the previous ones. However, they are quite complex, often requiring a model to verify its thought process by backtracking on faulty reasoning trajectories. As explained before, such requirements are difficult for smaller models to obey due to their limited instruction-following abilities. In fact, methods such as PAGER are seen to shine when models are scaled to 32B or 70B parameters. This is in contrast with our framework, which harnesses the strength of smaller models by avoiding heavy reasoning demands from them. 

Finally, we notice how well Standard RAG performs against all methods. This further motivates the need for developing simpler frameworks that are more suitable for the capabilities of SLMs.

\subsubsection{Information Efficiency}

\begin{table}[ht]
\centering
\small
\setlength{\tabcolsep}{2.8pt}
\begin{tabular}{l c c c}
\toprule
\textbf{Framework} & 
\makecell{\textbf{Avg. Context} \\ \textbf{Tokens (Payload)}} & 
\makecell{\textbf{Relative} \\ \textbf{Overhead}} & 
\makecell{\textbf{EM/F1}} \\
\midrule
HGRAG & \textasciitilde17K          & \textasciitilde13K\%  & 0.47 / 10.59  \\
Std. RAG   & 141.33             & 6.60\%          & 17.27 / 24.71 \\
Search-o1 & 141.32 & 6.54\% & 12.33 / 20.32  \\
GA   & 135.98             & 2.52\%          & 17.87 / 24.10 \\
\midrule
\textbf{Ours} & \textbf{132.64}    & - & \textbf{20.20 / 27.63} \\
\bottomrule
\end{tabular}
\vspace{-1.5mm}
\caption{Comparing average retrieved context (payload) tokens across all datasets. Relative Overhead indicates percentage increase in token consumption over our method. HGRAG (HyperGraphRAG), GA (GraphAnchor), Std. RAG (Standard RAG).}
\label{tab:token_comparison}
\vspace{-10pt}
\end{table}


To understand the computational demands for each method, we measure the average number of tokens in the \textit{retrieved contexts}. We do this to (i) gauge the core retrieval operation, i.e., the amount of grounding context needed for a framework to perform as designed, and (ii) strip out the variation due to prompt design, which can be succinct or verbose, and output token limits that can be controlled as a hyperparameter.

\begin{table*}[ht]
\centering
\newcolumntype{Y}{>{\centering\arraybackslash}X}
\newcommand{\scol}[1]{>{\centering\arraybackslash}S[table-format=#1, table-column-width=0.08\linewidth]}
\small
\setlength{\tabcolsep}{0pt}
\begin{tabularx}{\textwidth}{l *{8}{Y}}
\toprule
& \multicolumn{2}{c}{2WikiMultiHopQA} & \multicolumn{2}{c}{HotpotQA} & \multicolumn{2}{c}{MuSiQue} & \multicolumn{2}{c}{Average} \\
\cmidrule(lr){2-3} \cmidrule(lr){4-5} \cmidrule(lr){6-7} \cmidrule(lr){8-9}
Setting & {EM} & {F1} & {EM} & {F1} & {EM} & {F1} & {EM} & {F1} \\
\midrule
\textbf{Full Framework} & 29.60 & 34.71 & 20.80 & 28.75 & 10.20 & 19.42 & \textbf{20.20} & \textbf{27.63} \\
\midrule
\quad $-$ System-II           & 25.40 & 29.73 & 14.20 & 21.35 & 2.60  & 10.03 & 14.07\decr{-30.3} & 20.37\decr{-26.3} \\
\quad $-$ Initial Reasoning & 23.00 & 30.63 & 13.00 & 21.82 & 7.20  & 16.32 & 14.40\decr{-28.7} & 22.92\decr{-17} \\
\quad $+$ Threshold        & 30.40 & 35.47 & 18.13 & 25.52 & 7.20  & 15.25 & 18.58\decr{-8} & 25.41\decr{-8} \\
\bottomrule
\end{tabularx}
\vspace{-1.5mm}
\caption{Investigating impact of framework components (ablation). ``$-$ System-II'' indicates only System-I answering, i.e., without involving any reasoning/retrieval. ``$-$ Initial Reasoning'' components refer to answering without looking at the System-I answer and initial reasoning triples. ``$+$ Threshold'' uses confidence scores from the SLM to reduce the number of questions passing on to System-II processing.}
\label{tab:ablation}
\vspace{-5pt}
\end{table*}

The results are shown in Table \ref{tab:token_comparison}. As we can see, \textbf{the average payload (retrieved context) for our method is the lightest}, i.e., the least number of tokens. This is in stark contrast with HyperGraphRAG, which bloats context with hyperedges, leading to unmanageable API costs if used in production. Search-o1 and GraphAnchor also consume more tokens than ours, as both perform iterative retrieval across steps to answer questions. We are unable to measure PAGER, PruneRAG, and RT-RAG as PAGER has many questions that do not undergo retrieval, while the others do not provide a way to parse the collected chunks.

\subsection{RQ2: Analyzing Framework Components}


In Table \ref{tab:ablation}, we present the results of our investigation of the different facets of our strategy. Here, we ask: (i) How well does System-I alone work? (ii) What happens if the retrieved context is irrelevant and the model has no fall-back option? (iii) Is there a way to gauge model \textit{confidence} to balance System-I/II thinking?


\subsubsection{Removing System-II}

Firstly, we see that System-I alone (row 2), i.e., zero-shot, can provide reasonable accuracy. This makes sense, as even at 7B parameters, SLMs have sufficient parametric knowledge to at least make educated guesses. However, we do observe a drastic decline when compared with the full pipeline, highlighting the importance of grounding context to steer the model towards the correct answer.

\subsubsection{Excluding Initial Reasoning}

Next, we consider the impact of the System-I hypothesis and its associated reasoning (generated triples) on the final answer. When they are absent, the model does not have an answer to fall back on or understand where it went wrong when comparing the retrieved evidence with its generated trace (triples). This is reflected by reduced performance in row 3, highlighting their importance, where, if the retrieved context is inconclusive, the model can still rely on its initial hypothesis as an answer.

\subsubsection{Balancing System-I v/s II}
\label{sec:threshold}

In addition to the main framework, we were interested to see if there is a way to \textit{balance System-I/II}, i.e., decide which questions are complete, from System-I, and which need further processing (System-II). We do this for two reasons: (i) System-II processing of questions with already correct answers would be redundant, and (ii) noise in the retrieved context can sway the SLM from its initial answer, if it is already correct. To implement this, we utilize \textbf{model confidence}, i.e., answer logits. Confidence scores are calculated by a standard mean of log-softmax probabilities for the decoded answer tokens. Questions whose answer confidence is above an empirically determined threshold are marked as complete, with the remaining proceeding to System-II. Prior work \cite{mahaut-etal-2024-factual} explains that logit scores are an imperfect measure of a model's confidence as they reflect token probability over the actual content of the output. However, we treat them as a \textit{rough signal} to avoid the overhead of training a separate verification model \cite{zhou2025variation}. 




\begin{figure*}
    \centering
    \includegraphics[width=\linewidth]{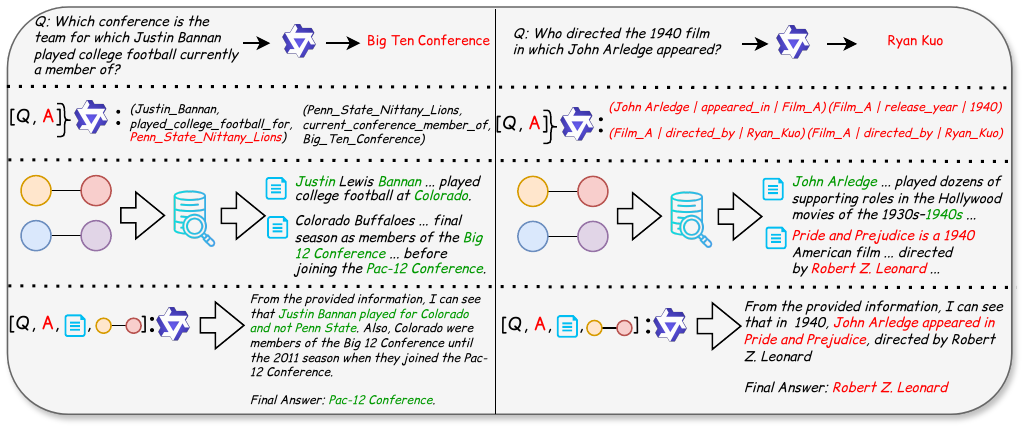}
    \vskip -0.5em
    \caption{Success/failure (left/right) cases of our framework. Further examples contrasting with others in App. \ref{sec:app_compare_rag}.}
    \label{fig:analysis}
    \vskip -0.5em
\end{figure*}

We notice that in general, model confidence for the majority of answers is quite low. This makes sense, as the questions require chaining interconnected pieces of information, and we are asking the model to answer without any form of thinking. Naturally, while it might know the answer, from its parametric knowledge, it is not confident in it. This is similar to how humans behave when asked a question that we potentially know the answer to, but lack the assertiveness in our judgment \cite{xie2026tip, yang2019feeling}. Thus, we set our threshold, as empirically observed, to .05, which yielded the best performance overall. Under this value, we notice that the majority of questions move to System-II. However, around 200 questions per dataset were marked complete, indicating a sizable proportion (App. \ref{sec:app_balancing_details}). Row 4 shows the results for this experiment. As we can see, even a lightweight filter can deliver performance close to the full pipeline but by being more efficient (App. \ref{sec:app_balancing_details}). This is beneficial in settings where the compute budget needs to be managed carefully.

\subsubsection{Quantifying System-I/II Advantage}

Finally, we study the net overall effect System-II reasoning has over System-I. For this, we calculate the number of questions whose answer changes between the two modes, either for the better or worse. Results are shown in Table \ref{tab:wrong_right_analysis}. We compute the number of questions whose answers are correct in either mode, going from being incorrect in System-I to correct in System-II (gain) and vice versa (loss). Correctness is measured using the same strict exact match metric used for the rest of our results. As such, the absolute numbers are generally low. Overall, we notice a \textbf{net positive impact of System-II} reasoning with the proposed framework, especially for MuSiQue, which is claimed to be more difficult than the other two \cite{trivedi-etal-2022-musique}.

\begin{table}[ht]
\centering
\small
\setlength{\tabcolsep}{4.4pt}
\begin{tabular}{l cccc}
\toprule
\textbf{Dataset} & \makecell{Both Correct\\($C \to C$)} & \makecell{Gained\\($I \to C$)} & \makecell{Lost\\($C \to I$)} & \makecell{\textbf{Net Gain}\\($\Delta$)} \\
\midrule
2Wiki & 82 & 66 & 45 & \cellcolor{green!15}21 \\
HTPT  & 49 & 55 & 22 & \cellcolor{green!15}33 \\
MUS   & 9  & 42 & 4  & \cellcolor{green!15}38 \\
\bottomrule
\end{tabular}
\vspace{-1.5mm}
\caption{Answer changes between System-I v/s System-II processing. 2wiki = 2WikiMultiHopQA, HTPT = HotpotQA, MUS = MuSiQue, $C$/$I$ = Correct/Incorrect. $\Delta = (Gain - Lost)$.}
\label{tab:wrong_right_analysis}
\end{table}

\subsection{RQ3: Qualitative Analysis}
\label{sec:qualitative_analysis}

Lastly, we perform a qualitative analysis of success and failure cases to understand the conditions under which the framework excels or struggles. We study two examples from \texttt{HotpotQA}, as shown in Fig. \ref{fig:analysis}.

On the left, our system works as expected. The model produces an incorrect, but plausible answer. However, from its generated triples, we learn why it believes \textit{Bannan} could be connected to the \textit{Big Ten Conference}, as it incorrectly reasons that he played for \textit{Penn State}. That said, these triples are useful anchors for locating counterfactual evidence. Using the first triple, the retriever finds an excerpt detailing where he played college football. Although the second triple does not contain direct links to \textit{Colorado football}, it describes elements such as team names, conferences, etc. that help the retriever locate top-k relevant items to it, out of which is an excerpt detailing the transition of the team to the \textit{Pac-12 conference}. By utilizing this counter-evidence for its initial claim, the model realizes its error and resolves to the correct answer.

The right side showcases an example where our framework struggles. When the model is \textit{truly uncertain} in its initial answer, its reasoning trace gets reflected accordingly. Here, the SLM is unable to recall which \textit{1940 film} the actor starred in. As such, it uses ineffective placeholder variables to justify its reasoning, which in turn leads to the wrong context being retrieved. Based on the retrieved context, the SLM reconciles the wrong \textit{1940 film} with the actor, ultimately producing a flawed answer.

Overall, we conclude that when the model is reasonably confident in its initial answer, its subsequent reasoning will drive it towards the correct answer. However, in cases where it simply does not have enough knowledge to make an educated guess, it will try its best to generate a faithful reasoning chain, which can ultimately harm its ability to find the true answer. As such, efforts at detecting model confidence (\S \ref{sec:threshold}) can make a difference here in building a more robust framework.

\section{Conclusion}

In this paper, we propose a cognitively inspired KB-TTS method, where we show that model \textit{hallucinations can be beneficial} as an initial clue to locate the real answer. We show that, against prior work, which requires multiple rounds of reasoning/retrieval, our method recognizes the limited capabilities of SLMs but uses that to our advantage. However, as seen in Sec. \ref{sec:qualitative_analysis}, future work needs to balance cases where the SLM truly struggles with an initial answer with a more deliberate stepwise reasoning approach (Fig. \ref{fig:motivation}) for a best-of-both-worlds setting.

\section*{Limitations}

We identify two limitations with our work: (i) The datasets that we use, like all MHQA studies, are factoid-style (requiring short and objective answers) questions based on general-domain topics. As such, it remains to be seen how well our method works across more complex domains such as medical or finance. Future work will aim to address these spaces; and (ii) Our method does not perform iterative reasoning/retrieval, which, in cases such as Fig. \ref{fig:analysis}, can be useful to mitigate incorrect reasoning trajectories. Although this does not impact the overall system performance, as observed empirically (Table \ref{tab:main_results}), future work will explore how to trade off zero-shot retrieval with a multi-stage pipeline to achieve even more robust performance.

\bibliography{references}

\appendix

\section{Balancing System-I/II Details}
\label{sec:app_balancing_details}

Table \ref{tab:balancing_stats} provides the number of samples that were marked ``complete'' according to the model confidence scores, and those that were deemed necessary for further System-II processing. As we can see, around \textasciitilde43\% ((238+185+227)/1500) of the questions were marked complete, indicating a sizable portion of samples not requiring further processing, which could end up being unnecessary.

To estimate how much compute would be required to process the completed questions, we calculate the average number of input tokens for each of those samples. This includes the prompts along with the retrieved context. As we can see, for each dataset, we save a considerable amount of prefill compute \cite{10.1145/3676641.3715996}, i.e., the pre-decoding phase where the entire input prompt is processed. Thus, if the compute budget is constrained, we can still see reasonable performance (Table \ref{tab:ablation}), but with a lower number of SLM calls.

\begin{table}[ht]
\centering
\small
\setlength{\tabcolsep}{3pt}
\begin{tabular}{lrrr}
\toprule
\textbf{Dataset} & \textbf{Complete} & \textbf{Incomplete} & \textbf{Avg. I/P Tokens Saved} \\ \midrule
2Wiki  & 238               & 262                 & 389.66                     \\
HTPT         & 185               & 315                 & 394.36                     \\
MUS          & 227               & 273                 & 393.17                     \\ \bottomrule
\end{tabular}
\caption{Number of questions marked ``complete'' from System-I and those that required further processing in System-II (incomplete). Avg. Tokens Saved = Average input (I/P) tokens for each ``complete'' question, if they had to go to System-II. 2wiki = 2WikiMultiHopQA, HTPT = HotpotQA, MUS = MuSiQue.}
\label{tab:balancing_stats}
\end{table}

\section{SLM Reasoning}
\label{sec:app_slm_reasoning}

To further lend credence to our \textit{answer-then-reason} hypothesis, we provide empirical justification. For the example question in Figure \ref{fig:motivation}, i.e., \textit{Which major Assasins Creed (AC) game is set in the city of London?}, we ask Qwen-2.5-7B, the same model used in all our experiments, to (i) perform step-by-step reasoning, shown in Figure \ref{fig:qwen_reasoning_for_example_question.} and, (ii) give a direct, System-I style answer in Figure \ref{fig:qwen_sys_I}.

As we can see, the \textit{reason-first} strategy indeed forces the model to search through all the games, to determine which title is set in \textit{London}. In theory, this seems reasonable. However, this will lead to at least nine searches (based on the chronological order of the games) to reach \textit{Syndicate}, causing major context bloat, which in turn, can lead to the model getting confused about which portion to pay attention to \cite{liu-etal-2024-lost}. On the other hand, our \textit{answer-first} strategy sees the model, quickly, albeit incorrectly, say \textit{Unity}. We use this as a search key to locate counterfactual information and, in turn, find out from its context that its successor, i.e., \textit{Syndicate}, is actually the title set in London. As such, our method can be considered as a \textit{backtracking} framework where search can be done once we start with a reasonable, educated guess.

\begin{figure}
    \centering
    \begin{llmreasoning}[title=Reason First]
        \textit{To solve this question, follow these steps:}

        \begin{enumerate}
            \item **\textbf{Identify the Assassin's Creed Games}**: List all the major Assassin's Creed games released so far.

            \item **\textbf{Research Each Game’s Setting}**: For each game on your list, find out where the game takes place.

            \item **\textbf{Identify the Game Set in London}**: From the research done in step 2, identify which game is set in the city of London.

            \item **\textbf{Provide the Answer}**: Once you have identified the correct game, provide the answer to the question.
        \end{enumerate}

    \textit{Following these steps will lead you to the correct answer.}
    \end{llmreasoning}
    \caption{Step-by-step reasoning for Fig. \ref{fig:motivation} question.}
    \label{fig:qwen_reasoning_for_example_question.}
\end{figure}

\begin{figure}
    \centering
    \begin{llmreasoning}[title=Answer First]
        \textit{The major Assassins Creed game set in the city of London is "Assassin's Creed IV: Black Flag" with a significant side mission set in London. However,} \textbf{the primary game set in London is "Assassin's Creed: Unity," which was released in 2014}
    \end{llmreasoning}
    \caption{System-I reasoning for Fig. \ref{fig:motivation} question.}
    \label{fig:qwen_sys_I}
\end{figure}

\section{Baseline Details}
\label{sec:app_baselines}

We provide further details on each baseline as follows,

\begin{enumerate}
    \item \textbf{Standard RAG} \cite{NEURIPS2020_6b493230}: This is the original RAG method that is used as a fixed baseline in any related study. The idea is simple: given a question, retrieve the top-k documents from a knowledge base that are relevant to it. These documents are included in the prompt for the model (LLM/SLM) as grounding context for the question. While this showed great promise initially in mitigating hallucinations and providing better answers, its performance was tested when subjected to more complex multi-step reasoning questions, as all of the necessary documents cannot be retrieved at once.

    \item \textbf{HyperGraphRAG} \cite{luo2025hypergraphrag}: Inspired by GraphRAG \cite{10.1145/3777378}, this method constructs a \textit{hypergraph}, i.e., a graph where edges connect multiple vertices. The idea behind doing this was to create a knowledge graph that broadly covers as much information as possible. However, in our experiments, we noticed that this method is incredibly inefficient. For instance, it is the only framework for which we had to use the gold contexts, as without it, index construction becomes infeasible. Even with the gold contexts, it takes 4 days to complete all three datasets, and without it, around 40 hours per dataset, which, if using an API-hosted model, incurs massive costs. This is confirmed by the authors: \url{https://github.com/LHRLAB/HyperGraphRAG/issues/14}.

    \item \textbf{GraphAnchor} \cite{liu2026graph}: Another variant of GraphRAG. Here, the framework uses an \textit{evolving} knowledge-graph, i.e., at each step, a small, focused knowledge-graph, based on the retrieved documents, is developed, which keeps track of the entities and relationships in them. This graph is used to augment the retrieved documents and provide a relational understanding of the data. Unfortunately, the complexity of firing multiple follow-up questions, developing the graph, and keeping track of the reasoning chain, prevented the model from even completing all of the questions (43\% across all datasets were incomplete). If we had used their evaluation script to report scores, each dataset would have close to \textasciitilde2\% F1, which would have heavily penalized their method. Thus, to be consistent with other methods, we use the official HuggingFace library to evaluate them, which reports appropriate numbers (App. \ref{sec:app_eval_dets}).

    \item \textbf{PruneRAG} \cite{jiao2026prunerag}: Grounded in  \textit{tree-based reasoning} \cite{NEURIPS2023_271db992}, i.e., exploring multiple search spaces to determine an answer, PruneRAG performs three operations: (i) query decomposition and tree construction, based on a model's reasoning using retrieved context; (ii) backtracking on the branches to aggregate evidence, (iii) token-confidence based answer rejection. While all of this provides decent performance, their method hinges on (i) question decomposition abilities of the model and (ii) token confidence, which, as explained \cite{mahaut-etal-2024-factual}, is not a reliable indicator of answer correctness.

    \item \textbf{RT-RAG} (Reasoning Tree Guided RAG) \cite{shi2026reasoning}: Similar to PruneRAG, RT-RAG also performs tree-style reasoning. It generates multiple candidate trees, selects the ``optimal'' one based on tree statistics such as depth and number of nodes, and, finally, gathers evidence based on the optimal tree to provide an answer.

    \item \textbf{PAGER} \cite{li2026structured}: This method first creates a ``cognitive outline'' (called ``page'') of what pieces of information a model needs to answer a question. The page has ``slots'' which are placeholders for potentially required data. These slots are progressively filled with iterative retrieval. Finally, once the entire page has been populated, the model provides its answer.

    \item \textbf{Search-o1} \cite{li-etal-2025-search}: Inspired by \texttt{Agentic RAG} \cite{singh2025agentic}, i.e., letting an LLM-agent dictate what to retrieve instead of providing it beforehand with top-k documents. Search-o1 differs from the others in two ways: (i) performing \textit{web} searches to retrieve background data, instead of using a pre-defined knowledge index, and (ii) including a module to refine the retrieved web pages to only provide the agent with the most relevant context for the question. We use the implementation provided by \citet{jiao2026prunerag} as they modify it to use a local index rather than web searches.
\end{enumerate}

\section{Prompts}

Here we provide the prompts used in our framework in Figures \ref{fig:sys1_prompt}, \ref{fig:triple_gen_prompt}, \ref{fig:sys2_prompt}.

\begin{figure}[htbp]
    \centering
    \begin{tcolorbox}[promptbox={Prompt For System-I Answering}, width=\columnwidth] 
    You are a precise answering engine. Your task is to provide the direct answer to a question without any explanation.

\#\#\# Rules:
\begin{enumerate}
    \item Provide ONLY the specific answer. 
    \item Do not include introductory phrases (e.g., "The answer is..."), explanations, or context.
    \item The answer must be wrapped in <answer> tags inside an <output> block.
    \item Output ONLY the <output> block.
\end{enumerate}

\#\#\# Example:\\
<input>\\
Question: What is the capital of France?
</input>\\

<output>\\
<answer>Paris</answer>\\
</output>\\

\#\#\# Task:
Process the following input and provide the answer within an <output> block.
    \end{tcolorbox}

    \caption{System-I answer prompt.}
    \label{fig:sys1_prompt}
\end{figure}

\begin{figure}[htbp]
    \centering
    \begin{tcolorbox}[promptbox={Prompt For Triple Generation}, width=\columnwidth] 
    You are a knowledge graph extractor. Your task is to generate a detailed logical sequence of subject-predicate-object triples that derive a given Answer from a given Question.

\#\#\# Rules:
\begin{enumerate}
    \item Format: Each triple must be enclosed in <triple> tags using the structure: <triple>Subject | predicate\_link | Object</triple>.
    \item Logical Depth: Do not skip steps. If the question involves a specific role or relationship (e.g., "X's lead singer" or "Y's director"), you MUST identify that specific individual as a separate node before linking them to the final answer.
    \item Chain of Reasoning: The sequence must form a step-by-step path where the Object of one triple leads to the Subject of the next.
    \item Predicate Style: Use concise, lowercase, snake\_case for predicates.
    \item Strict Output: Provide ONLY the <output> block. Do not include introductory text or explanations.    
\end{enumerate}

\#\#\# Examples:

<input>\\
Question: Where was the lead singer of the band Queen born?\\
Answer: Stone Town, Zanzibar\\
</input>

<output>\\
<triple>Queen | has\_lead\_singer | Freddie Mercury</triple>\\
<triple>Freddie Mercury | born\_in | Stone Town, Zanzibar</triple>\\
</output>\\

<input>\\
Question: What is the birthplace of the person who designed the Eiffel Tower?\\
Answer: Dijon, France\\
</input>

<output>\\
<triple>Eiffel Tower | designed\_by | Gustave Eiffel</triple>\\
<triple>Gustave Eiffel | born\_in | Dijon, France</triple>\\
</output>\\

\#\#\# Task:
Process the following input and provide the triples within an <output> block.
    \end{tcolorbox}

    \caption{Triple generation prompt.}
    \label{fig:triple_gen_prompt}
\end{figure}

\begin{figure}[htbp]
    \centering
    \begin{tcolorbox}[promptbox={Prompt For System-II Answering}, width=\columnwidth] 
    You are a question answering assistant. You are given a question, an initial guess, supporting evidence for that guess (as knowledge graph triples) and retrieved context related to the question. \\

Think about everything step-by-step, by considering all of the information and determining where the flaws are. \\

Provide a clear, structured explanation of your logic, and conclude by stating the final answer clearly. \\

Always wrap your final answer inside <final\_answer> [answer] </final\_answer> tags.
    \end{tcolorbox}

    \caption{System-II answer prompt.}
    \label{fig:sys2_prompt}
\end{figure}

\section{Code}

Our code is available at \url{https://github.com/saptarshi059/TTS-project}.

\section{Evaluation Details}
\label{sec:app_eval_dets}

All frameworks are evaluated using the HuggingFace implementation (\url{https://github.com/huggingface/evaluate}) of the metrics of the original work by \citet{rajpurkar-etal-2018-know}.

\section{Comparing RAG Methods}
\label{sec:app_compare_rag}

In Figures \ref{fig:std_rag}, \ref{fig:graph_anchor}, and \ref{fig:our_method}, we contrast vanilla RAG and GraphAnchor with our method to show where each method struggles and where we excel. 

Firstly, standard RAG pulls the top-5 most relevant documents for the query. Apart from the first document, which describes where \textit{Justin Bannan} played, the others are irrelevant. As such, the model does not find what it is looking for, resorting to its parametric knowledge, as nowhere is it mentioned that \textit{Colorado} played in the \textit{Big 12 Conference}.

GraphAnchor, in this example, is surprisingly worse than standard RAG. First, it retrieves the same documents as standard RAG. It correctly identifies that \textit{Justin Bannan} played for \textit{Colorado}. However, it likely believes that \textit{Colorado} has the \textit{Carroll Fighting Saints} team, based on the generated graph, and thus states that he played in the \textit{Frontier Conference}. This example highlights the limitations of GraphRAG approaches, where failure to understand relationships can lead to incorrect answers.

Finally, in our method, we observe a better coverage of information coming from two retrievals, based on the two generated triples. Using this, the SLM realizes that while \textit{Colorado} initially played in the \textit{Big 12 Conference}, they later joined \textit{Pac-12}. This overrules the initial guess (\textit{Big Ten Conference}), which turned out to be quite useful in generating the triples that led to the acquisition of the relevant documents.

\section{Hardware Requirements}

As we are dealing with small language models, our hardware demands are relatively modest. Our cluster gives us access to NVIDIA A100 80 GB GPUs. However, we use only a \textit{single GPU}, which loads both SLM and retriever models quite comfortably.

\begin{figure*}
    \begin{ragcomparison}[title={\textit{Q: Which conference is the team for which Justin Bannan played college football currently a member of?}}]

\methodsection{Standard RAG}{%
    Documents:

\begin{enumerate}[leftmargin=*]
    \item \textcolor{ForestGreen}{Justin Lewis Bannan (born April 18, 1979) is a former American football defensive tackle.  He was drafted by the Buffalo Bills in the fifth round of the 2002 NFL Draft.  He played college football at Colorado.}

    \item \textcolor{CadetBlue}{Justin Harper (born February 24, 1985 in Catawba, North Carolina) is an American college football coach and former professional Canadian football wide receiver ...}

    \item \textcolor{CadetBlue}{Timothy Joseph "Tim" Landis (born July 13, 1964) is an American football coach who is currently quarterbacks coach and special teams coordinator at Lycoming College ...}

    \item \textcolor{CadetBlue}{Benjamin "Bennie" Oosterbaan (February 4, 1906 – October 25, 1990) ...}

    \item \textcolor{CadetBlue}{The Carroll Fighting Saint football program represents Carroll College of Helena, Montana in college football ...}
\end{enumerate}

\brokenline

\textit{To answer this question, let's break down the information provided:}

\begin{enumerate}[leftmargin=*]
    \item \textit{Justin Bannan played college football at Colorado.}
    \item \textit{We need to find out which conference the current team representing Colorado is a member of.}
\end{enumerate}

\textit{From the given information, we can infer that the current team representing Colorado is part of the Big 12 Conference, as Colorado is a member of the Big 12 Conference.}\\

\textit{Therefore, the answer is the Big 12 Conference.}\\

\textcolor{red}{\textit{<final\_answer> Big 12 Conference </final\_answer>}}

}
\end{ragcomparison}
\caption{Demonstrating Standard RAG on an example question.}
\label{fig:std_rag}
\end{figure*}

\begin{figure*}
    \begin{ragcomparison}[title={\textit{Q: Which conference is the team for which Justin Bannan played college football currently a member of?}}]

\methodsection{GraphAnchor}{%
    Documents:

\begin{enumerate}[leftmargin=*]
    \itemsep0em 
    \item \textcolor{ForestGreen}{Justin Lewis Bannan (born April 18, 1979) is a former American football defensive tackle.  He was drafted by the Buffalo Bills in the fifth round of the 2002 NFL Draft.  He played college football at Colorado.}

    \item \textcolor{CadetBlue}{Justin Harper (born February 24, 1985 in Catawba, North Carolina) is an American college football coach and former professional Canadian football wide receiver ...}

    \item \textcolor{CadetBlue}{Timothy Joseph "Tim" Landis (born July 13, 1964) is an American football coach who is currently quarterbacks coach and special teams coordinator at Lycoming College ...}

    \item \textcolor{CadetBlue}{Benjamin "Bennie" Oosterbaan (February 4, 1906 – October 25, 1990) ...}

    \item \textcolor{CadetBlue}{The Carroll Fighting Saint football program represents Carroll College of Helena, Montana in college football ...}
\end{enumerate}

\brokenline

\textit{<think>}\\
\textit{Based on the question, we need to identify the current conference of the team for which Justin Bannan played college football. From the given documents, we can find the following information:}

\begin{itemize}[label=-, leftmargin=*]
    \itemsep0em 
    \item \textit{Justin Bannan played college football at Colorado.}
    \item \textit{The team for which Justin Bannan played college football is currently a member of the Frontier Conference.}
\end{itemize}

\textit{Therefore, we have sufficient information to answer the question.}\\
\textit{</think>}

\vspace{2pt}
\textit{<judgement> sufficient </judgement>}
\vspace{2pt}

\textit{<graph>}

\textit{Entities:}
\begin{itemize}[label=-]
    \itemsep0em 
    \item \textit{[Carroll Fighting Saints football] (Attributes: [current conference: Frontier Conference])}
    \item \textit{[Justin Bannan] (Attributes: [played college football at: Colorado])}
\end{itemize}

\textit{Relationships:}
\begin{enumerate}
    \itemsep0em 
    \item \textit{[Carroll Fighting Saints football] -> [current conference] -> [Frontier Conference]}
    \item \textit{[Justin Bannan] -> [played college football at] -> [Colorado]}
\end{enumerate}
\vspace{-\topsep}
\textit{</graph>}\\

\textit{\textcolor{red}{<final\_answer> Frontier Conference </final\_answer>}}

}
\end{ragcomparison}
\caption{Demonstrating GraphAnchor on an example question.}
\label{fig:graph_anchor}
\end{figure*}

\begin{figure*}
    \begin{ragcomparison}[title={\textit{Q: Which conference is the team for which Justin Bannan played college football currently a member of?}}]

\methodsection{Proposed Method}{%

Initial Guess: \textcolor{red}{Big Ten Conference} \\

Initial Reasoning: (Justin\_Bannan | played\_college\_football\_for | \textcolor{red}{Penn\_State\_Nittany\_Lions}), (Penn\_State\_Nittany\_Lions | current\_conference\_member\_of | Big\_Ten\_Conference) \\

Documents:

\begin{enumerate}[leftmargin=*]
    \itemsep0em 
    \item \textcolor{ForestGreen}{Justin Lewis Bannan (born April 18, 1979) is a former American football defensive tackle.  He was drafted by the Buffalo Bills in the fifth round of the 2002 NFL Draft.  He played college football at Colorado.}

    \item \textcolor{CadetBlue}{Benjamin "Bennie" Oosterbaan (February 4, 1906 – October 25, 1990) ...}

    \item \textcolor{CadetBlue}{Timothy Joseph "Tim" Landis ...}

    \item \textcolor{CadetBlue}{Justin Harper (born February 24, 1985 in Catawba, North Carolina) is ...}

    \item \textcolor{CadetBlue}{Judson Albert "Jud" Timm (August 28, 1906 – December 23, 1994) ...}

    \item \textcolor{CadetBlue}{The Trenton Lightning were an indoor ...}

    \item \textcolor{CadetBlue}{The 2013 Texas Longhorns football team ...}

    \item \textcolor{ForestGreen}{... Colorado Buffaloes ... It was also the final season as members the Big 12 Conference in the North Division for Colorado, before joining the Pac-12 Conference for the 2011 season ...}

    \item \textcolor{CadetBlue}{Roy Randolph Roundtree ...}

    \item \textcolor{CadetBlue}{The 2015 Wisconsin Badgers football ... }
\end{enumerate}

\brokenline

\textit{Let's break down the information step-by-step:}

\begin{enumerate}[leftmargin=*]
    \itemsep0em 
    \item \textit{**\textbf{Identify the Team}**: The question asks about the team for which Justin Bannan played college football. From the provided information, we know that Justin Bannan played college football for the Colorado Buffaloes.}

    \item \textit{**\textbf{Determine the Current Conference}**: The question asks about the current conference of the team for which Justin Bannan played college football. From the provided information, we know that the Colorado Buffaloes were members of the Big 12 Conference until the 2011 season when they joined the Pac-12 Conference.}
\end{enumerate}

\textit{Therefore, the current conference of the team for which Justin Bannan played college football is the Pac-12 Conference.} \\

\textcolor{ForestGreen}{\textit{<final\_answer> Pac-12 Conference </final\_answer>}}

}
\end{ragcomparison}
\caption{Demonstrating our method on an example question.}
\label{fig:our_method}
\end{figure*}

\end{document}